\documentclass{article} 
\usepackage{iclr2015,times}
\usepackage{hyperref}
\usepackage{url}
\usepackage{graphicx}
\usepackage{amsmath}
\usepackage{floatrow}
\usepackage{sidecap}
\usepackage{wrapfig}
\usepackage{subcaption}

\title{Gradual Training Method for Denoising Auto Encoders}

\author{
Alexander Kalmanovich \& Gal Chechik \\ 
The Gonda Brain Research Center\\
Bar Ilan University\\
52900 Ramat-Gan, Israel\\
\texttt{sashakal@gmail.com, gal.chechik@biu.ac.il} \\
}

\newcommand{\first}{1\textsuperscript{st}}
\newcommand{\second}{2\textsuperscript{nd}}
\newcommand{\third}{3\textsuperscript{rd}}
\newcommand{\ignore}[1]{} 
\iclrfinalcopy 

\begin{document}

\maketitle

\begin{abstract}
Stacked denoising auto encoders (DAEs) are well known  to learn useful deep representations, which can be used to improve supervised training by initializing a deep network. We investigate a training scheme of a deep DAE, where DAE layers are gradually added and keep adapting as additional layers are added. We show that in the regime of  mid-sized datasets, this \textbf{\textit{gradual training}} provides a small but consistent improvement over stacked training in both reconstruction quality and classification error over stacked training on MNIST and CIFAR datasets.
\end{abstract}

\section{Gradual training of denoising autoencoders}
We test here \textbf{\textit{‘gradual training’}} of deep denoising auto encoders, training the network layer-by-layer, but lower layers keep adapting throughout training. To allow lower layers to adapt continuously, noise is injected at the input level. This training procedure differs from stack-training of auto encoders ~\citep{vincent2010stacked}

More specifically, in gradual training, the first layer of the deep DAE is trained as in stacked training, producing a layer of weights $w_1$. Then, when adding the second layer autoencoder, its weights $w_2$ are tuned jointly with the already-trained weights $w_1$. Given a training sample $x$, we generate a noisy version $\tilde{x}$, feed it to the 2-layered DAE, and compute the activation at the subsequent layers
$h_1=Sigmoid(w^\top_1x)$, $h_2=Sigmoid(w^\top_2h_1)$ and
$y=Sigmoid({w_3'^\top}h_2)$. Importantly, the loss function is  computed over the input $x$, and is used to update all the weights including $w_1$. Similarly, if a \third{} layer is trained, it involves tuning $w_1$ and $w_2$ in addition to $w_3$ and $w'_4$.


\section{Experimental procedures}

We compare the performance of gradual and stacked training in two learning setups: an unsupervised denoising task, and a supervised classification task initialized using the weights learned in an unsupervised way. Evaluations were made on three benchmarks: MNIST, 
CIFAR-10 and CIFAR-100, 
but only show here MNIST results due to space constraints. We used a test subset of 10,000 samples and several sizes of training-set all maintaining the uniform distribution over classes. 

Hyper parameters were selected using a second level of cross validation, including the learning rate, SGD batch size, momentum and weight decay. In the supervised experiments, training was 'early stopped' after 35 epochs without improvement. The results reported below are averages over 3 train-validation splits. Since gradual training involves updating lower layers, every presentation of a sample involves more weight updates than in a single-layered DAE. To compare stacked and gradual training on a common ground, we limited gradual training to use the same ‘budget’ of weight update steps as stacked training. 

For example, when training the second layer for $n$ epochs in gradual training, we allocate $2n$ training epochs for stacked training (details in the full paper). 
\begin{figure}
\floatbox[{\capbeside\thisfloatsetup{capbesideposition={right,top},capbesidewidth=4cm}}]{figure}[\FBwidth]
{\caption{Unsupervised and supervised training results on MNIST dataset. Error bars are over 3 train-validation splits. Network has 2 hidden layers with 1000 units each \textbf{(a)} Reconstruction error of unsupervised training methods measured by cross-entropy loss. The shown cross-entropy error is relative to the minimum possible}
\label{figure1}}
{\includegraphics[clip, trim = 0.70cm 10cm 9.3cm 0cm,height=5.38cm] {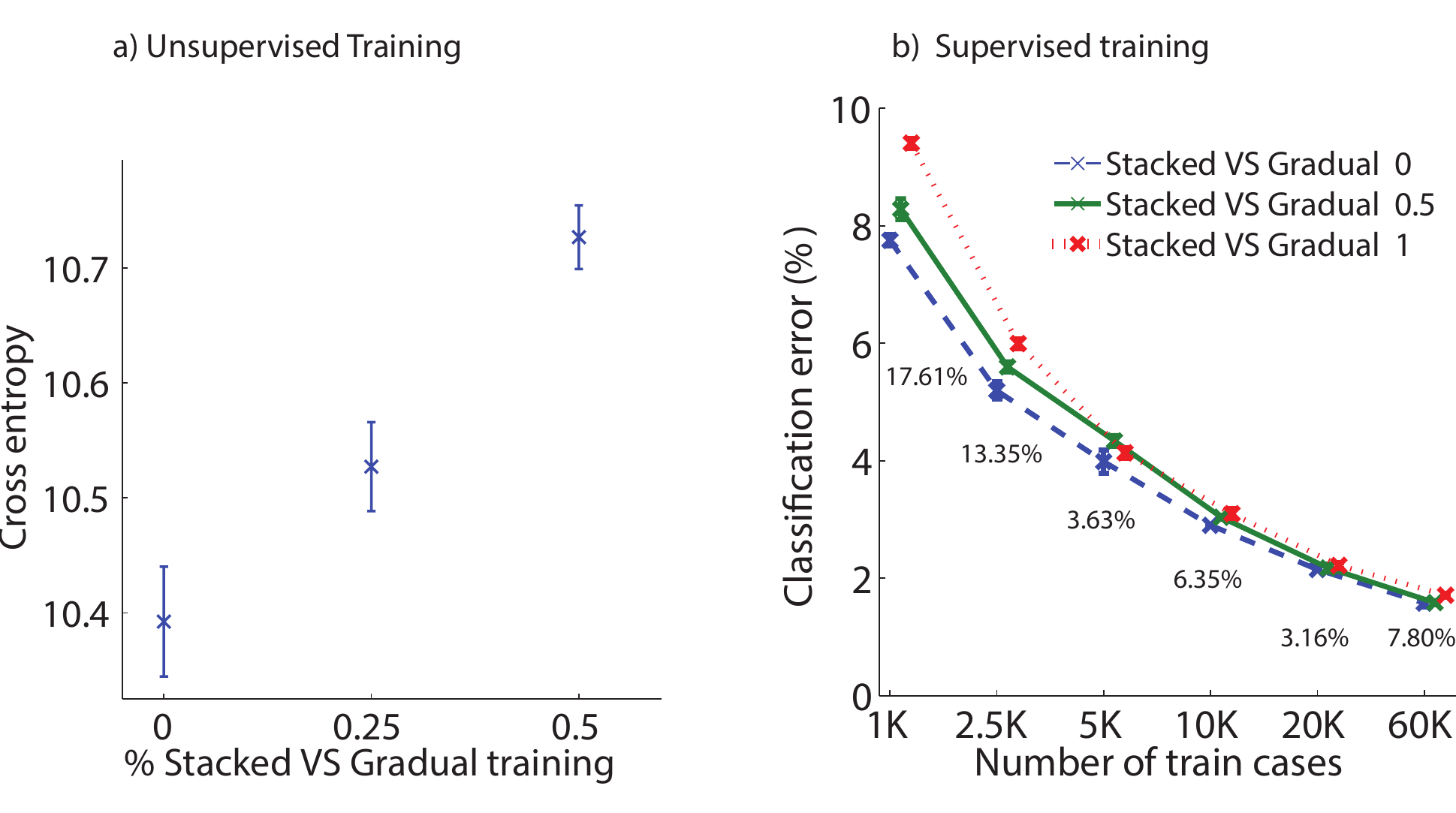}}
\end{figure}

\vspace{1.2cm}
error, computed as the cross-entropy error of the original uncorrupted test set with itself. All compared methods used the same budget of update operations. Images were corrupted with 15\% masking noise. The \first{} hidden layer is trained for 50 epochs. Total epoch budget for the \second{} hidden layer is 80 epochs.
\textbf{(b)} Classification error of supervised training initialized based on DAEs. Each curve shows a different pre-training type. Text labels show the percentage of error improvement of \textit{Stacked-vs-Gradual 0} pretraining compared to \textit{Stacked-vs-Gradual 1} pretraining.

\section{Results}
We evaluate gradual and stacked training in unsupervised task of image denoising, and then evaluated the quality of the two methods for initializing a network in a supervised learning task.
\ignore{
\begin{wrapfigure}{r}{0.70\textwidth}
\includegraphics[clip, trim = 0.5cm 10cm 9.3cm 0cm,height=5.38cm]{Fig1-SupAndUnsup.pdf}
\caption{Unsupervised and supervised training results on MNIST dataset. Error bars are over 3 train-validation splits. Network has 2 hidden layers with 1000 units each. \textbf{(a)} Reconstruction error of unsupervised training methods measured by cross-entropy loss. The shown cross-entropy error is relative to the minimum possible error, computed as the cross-entropy error of the original uncorrupted test set with itself. All compared methods used the same budget of update operations. Images were corrupted with 15\% masking noise. The \first{} hidden layer is trained for 50 epochs. Total epoch budget for the \second{} hidden layer is 80 epochs.
\textbf{(b)} Classification error of supervised training initialized based on DAEs. Each curve shows a different pre-training type. Text labels show the percentage of error improvement of \textit{Stacked-vs-Gradual 0} pretraining compared to \textit{Stacked-vs-Gradual 1} pretraining.}
\label{figure1}
\end{wrapfigure}
}

\textbf{Unsupervised learning for denoising}. 
We first evaluate gradual training in an unsupervised task of image denoising. Here, the network is trained to minimize a cross-entropy loss over corrupted images.   
In addition to stacked and gradual training, we also tested a hybrid method that spends some epochs on tuning only the second layer (as in stacked training), and then spends the rest of the training budget on both layers (as in gradual training). We define the \textit{Stacked-vs-Gradual} fraction $0\leq f \leq1$ as the fraction of weight updates that occur during ‘stacked’-type training. $f=1$ is equivalent to pure stacked training while $f=0$ is equivalent to pure gradual training. Given a budget of n training epochs, we train the \second{} hidden layer with gradual training for $n(1-f)$ epochs, and with stacked training for $2nf$ epochs. More specifically, since stacked training tunes a single layer of weights and gradual training tunes two layers of weights, we selected the number of stacked epochs $s$, and the number of gradual epochs $g$, such that $s+2g=n$, and changed several values of $s$ and $g$ to get different ratios \(f=1-\frac{g}{n}\).

Figure~\ref{figure1}a shows the test-set cross entropy error when training 2-layered DAEs, as a function of the \textit{Stacked-vs-Gradual} fraction. Pure gradual training achieved significant lower reconstruction error than any mix of stacked and gradual training with the same budget of update steps.
\ignore{
\begin{figure}[h]
\includegraphics[clip, trim = 0.5cm 10cm 9.3cm 0cm,height=5.38cm]{Fig1-SupAndUnsup.pdf}
\caption{Unsupervised and supervised training results on MNIST dataset. Error bars are over 3 train-validation splits. Network has 2 hidden layers with 1000 units each. \textbf{(a)} Reconstruction error of unsupervised training methods measured by cross-entropy loss. The shown cross-entropy error is relative to the minimum possible error, computed as the cross-entropy error of the original uncorrupted test set with itself. All compared methods used the same budget of update operations. Images were corrupted with 15\% masking noise. The \first{} hidden layer is trained for 50 epochs. Total epoch budget for the \second{} hidden layer is 80 epochs.
\textbf{(b)} Classification error of supervised training initialized based on DAEs. Each curve shows a different pre-training type. Text labels show the percentage of error improvement of \textit{Stacked-vs-Gradual 0} pretraining compared to \textit{Stacked-vs-Gradual 1} pretraining.}
\label{figure1}
\end{figure}
}

\textbf{Gradual-training DAE for initializing a network in a supervised task}. We further tested the benefits of using the DAEs trained in the previous experiment for initializing a deep network in a supervised classification task. We initialized the first two layers of the deep network with the weights of the SDAE and added a classification layer on top with output units matching the classes in the dataset, with randomly initialized weights. 

To quantify the benefit of gradual unsupervised pretraining we trained these networks on subsets of the training set. Figure.~\ref{figure1}b traces the classification error as a function of training set size, demonstrating a consistent but small improvement when using gradual training over stacked training (text legends). This effect is mostly relevant for datasets with less than $50K$ samples. Similar results were obtained using CIFAR10 and CIFAR100.

\bibliography{iclr2015}
\bibliographystyle{iclr2015}

\end{document}